\documentclass[letterpaper, 10 pt, conference]{cls/ieeeconf}
\IEEEoverridecommandlockouts
\let\labelindent\relax

\usepackage[normalem]{ulem}                         
\usepackage[table,usenames,dvipsnames]{xcolor}      
\usepackage{enumitem}                               
\usepackage{extarrows}                              
\usepackage[noadjust]{cite}                         
\usepackage{verbatim}

\usepackage{amsmath,amssymb,amsfonts,dsfont} 
\usepackage{amsthm}
\usepackage{bbm}
\usepackage{algorithm,algorithmicx,listings}        
\usepackage[noend]{algpseudocode}             

\usepackage{graphicx,tabularx,subcaption}
\usepackage[export]{adjustbox}
\usepackage{makecell,booktabs}

\usepackage[font={small}]{caption}
\usepackage[titletoc,title]{appendix}

\usepackage[breaklinks=true, colorlinks, bookmarks=true, citecolor=Black, urlcolor=Violet,linkcolor=Black]{hyperref}

\newcommand{\prl}[1]{\left(#1\right)}
\newcommand{\brl}[1]{\left[#1\right]}
\newcommand{\crl}[1]{\left\{#1\right\}}

\newcommand\norm[1]{\lVert#1\rVert}
\usepackage{mathtools}
\usepackage{pifont}
\usepackage{footnote}
\usepackage{tikz}
\usepackage{balance}
\makesavenoteenv{tabular}
\makesavenoteenv{table}

\def\etal/{et~al.}
\graphicspath{{fig/}}

\theoremstyle{definition}

\newtheorem*{assumption*}{Assumption}
\newtheorem*{problem*}{Problem}

\theoremstyle{remark}

\newtheorem*{solution*}{Solution}

\def\thetitle{Cross-Embodiment Robot Manipulation Skill Transfer using Latent Space Alignment}
\def\theauthor{Tianyu Wang and Dwait Bhatt and Xiaolong Wang and Nikolay Atanasov}
\def\thekeywords{keywords}

\hypersetup{
  pdfauthor={\theauthor},%
  pdftitle={\thetitle},%
  pdfsubject={\thetitle},%
  pdfkeywords={\thekeywords}
}

\newcommand{\calA}{{\cal A}}

\newcommand{\calD}{{\cal D}}

\newcommand{\calL}{{\cal L}}
\newcommand{\calM}{{\cal M}}
\newcommand{\calN}{{\cal N}}

\newcommand{\calS}{{\cal S}}

\newcommand{\bfa}{\mathbf{a}}

\newcommand{\bfs}{\mathbf{s}}

\newcommand{\bbE}{\mathbb{E}}

\newcommand{\bbR}{\mathbb{R}}

\newcommand{\tdF}{\tilde{F}}
\newcommand{\tdG}{\tilde{G}}
\newcommand{\tdT}{\tilde{T}}
\newcommand{\brbfs}{\bar{\bfs}}
\newcommand{\brbfa}{\bar{\bfa}}

\title{\LARGE \bf \thetitle}
\author{Tianyu Wang \and Dwait Bhatt \and Xiaolong Wang \and Nikolay Atanasov
\thanks{This work was supported by the Technology Innovation Program 20018112 (Development of autonomous manipulation and gripping technology using imitation learning based on visual and tactile sensing) funded by the Ministry of Trade, Industry \& Energy (MOTIE), Korea. The authors are with the Department of Electrical and Computer Engineering, University of California San Diego, La Jolla, CA 92093, USA (e-mails: {\tt\small \{tiw161,dhbhatt,xiw012,natanasov\}@ucsd.edu}).}%
}

\begin{document}
\maketitle


\begin{abstract}
This paper focuses on transferring control policies between robot manipulators with different morphology. While reinforcement learning (RL) methods have shown successful results in robot manipulation tasks, transferring a trained policy from simulation to a real robot or deploying it on a robot with different states, actions, or kinematics is challenging. To achieve cross-embodiment policy transfer, our key insight is to project the state and action spaces of the source and target robots to a common latent space representation. We first introduce encoders and decoders to associate the states and actions of the source robot with a latent space. The encoders, decoders, and a latent space control policy are trained simultaneously using loss functions measuring task performance, latent dynamics consistency, and encoder-decoder ability to reconstruct the original states and actions. To transfer the learned control policy, we only need to train target encoders and decoders that align a new target domain to the latent space. We use generative adversarial training with cycle consistency and latent dynamics losses without access to the task reward or reward tuning in the target domain. We demonstrate sim-to-sim and sim-to-real manipulation policy transfer with source and target robots of different states, actions, and embodiments. The source code is available at \url{https://github.com/ExistentialRobotics/cross_embodiment_transfer}.
\end{abstract}


\section{Introduction}

Reinforcement learning (RL) has achieved remarkable success in solving sequential decision-making problems where an agent improves its performance through interactions with the environment. Despite its success, one challenge that prevents the real-world application of RL is sample efficiency. RL training typically requires millions of interactions to obtain a policy specialized for a single task. On the other hand, humans exhibit the ability to learn from third-person observations of different embodiments \cite{Rizzolatti2004Mirror, Iacoboni1999Cortical, Stadie2017Third}. For example, if a human is able to pour a cup of tea with their right hand, they should need little practice to do it with their left hand. However, if a robot is trained with RL to pick an object up with one arm, the learned policy cannot be easily reused if the arm joint locations, link lengths, or arm dynamics chance.

\begin{figure}[t]
    \centering
    \includegraphics[width=\linewidth]{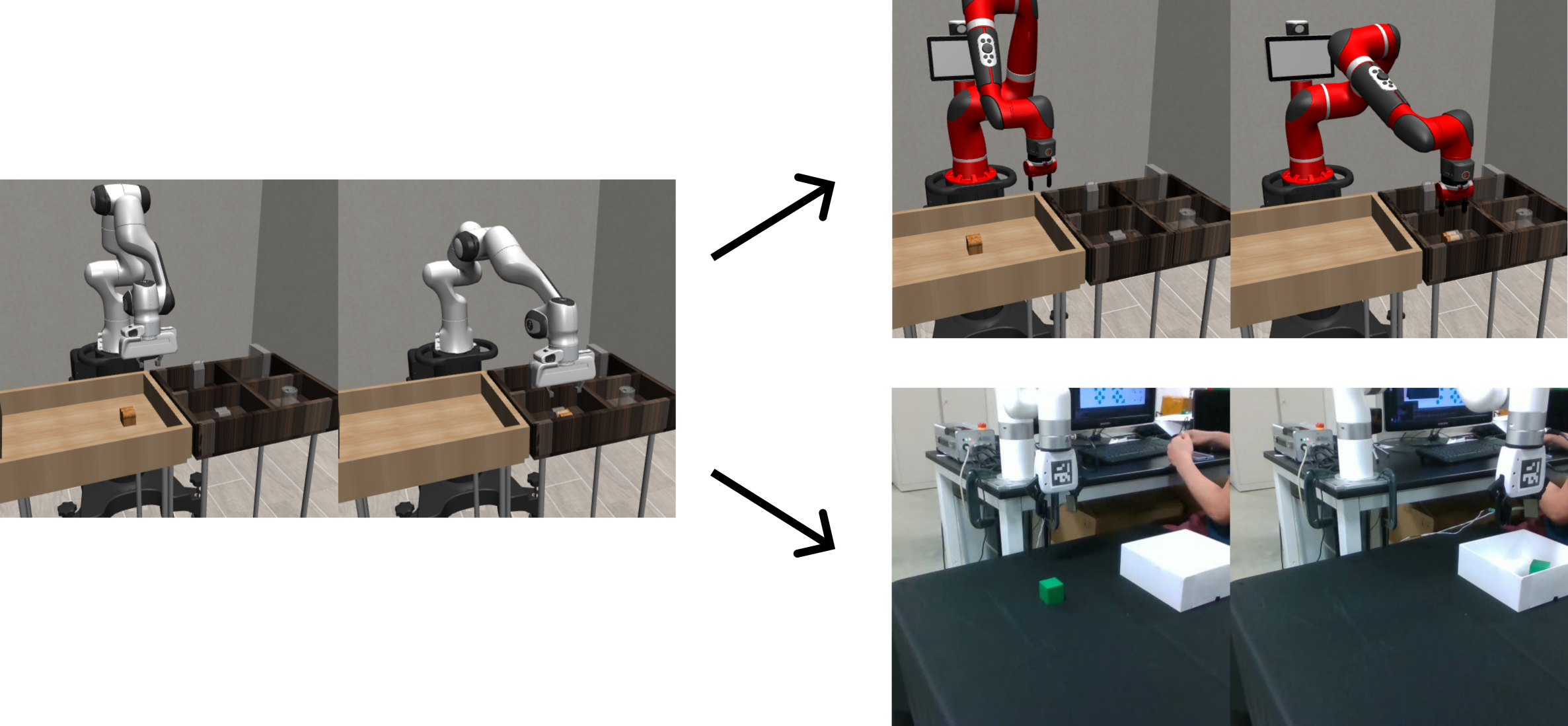}
    \caption{Policy transfer to different robot embodiments using latent space alignment. In the source domain (left), we train a simulated Panda robot to pick and place objects. Our approach allows transferring the policy to different target domains (right), such as a simulated Sawyer robot (top right) or a real xArm6 robot (bottom right), without requiring additional task-specific training data.}
    \label{fig:intro_transfer}
\end{figure}

Transfer learning is a promising methodology to reuse robot skills in different domains. This paper considers transfer learning between robots of different morphologies executing the same task (see Fig~\ref{fig:intro_transfer}). Recently, many works have proposed solutions for domain discrepancies using paired and aligned demonstrations \cite{Gupta2017Learning, Zhang2018Deep, Florence2019Self, Sermanet2018Time}. For example, temporal alignment assumes different robots can solve the same task at roughly the same speed. However, paired trajectories collected by a pretrained policy or human labelling are challenging to obtain. In self-supervised learning, latent representations are learned from pixel observations for prediction and reasoning in downstream supervised learning and RL applications. This is usually done with autoencoders using pixel reconstruction losses \cite{Watter2015E2C, Ha2018World, Hafner2019PlaNet} or contrastive learning \cite{Oord2018CPC, Eysenbach2022Contrastive, Anand2019Unsupervised} with energy-based losses \cite{Lecun2006Tutorial}. However, pixel reconstruction aims at decoding the original observations accurately, which may ignore visually small but important features, such as bullets in Atari games \cite{Kaiser2019Model}, or could focus on predicting irrelevant features, such as background. In contrastive learning, the number of samples needed to construct a well-shaped energy surface may scale exponentially, requiring significant computation to train the system \cite{Lecun2022Path}.



\begin{figure*}[t]
\centering
\includegraphics[width=0.95\linewidth]{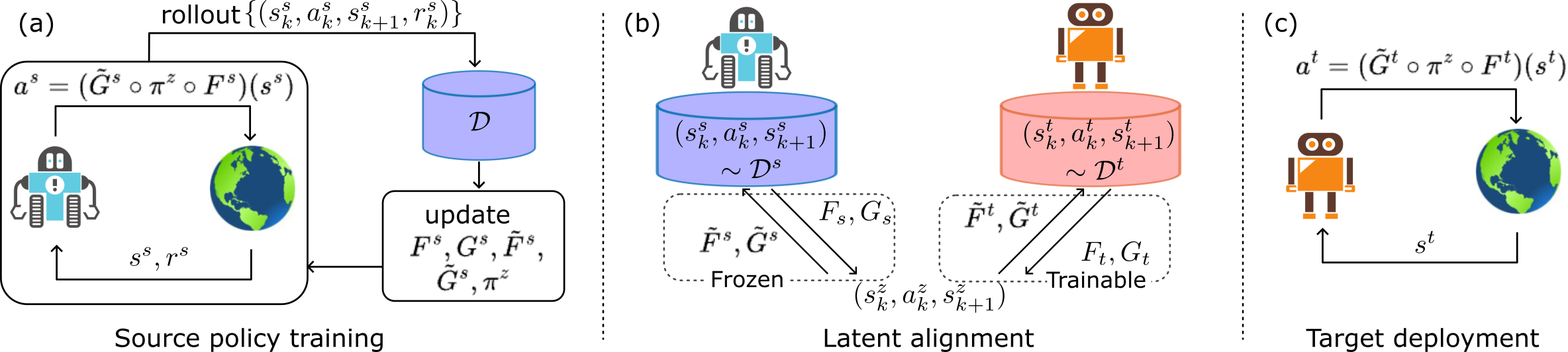}
\caption{Approach overview: (a) The source robot learns encoders and decoders $F_s, G_s, \tdF_s, \tdG_s$ for state-action projections between its own space and a latent space. The source robot learns a latent policy $\pi^z$ simultaneously with encoders and decoders with RL. (b) During latent alignment, the source encoder decoder functions are frozen while the target encoder decoder are trained to match latent distributions as well as to satisfy cycle consistency and latent dynamics constraints. (c) During target deployment, we compose the target encoder and decoder functions trained in (b) with the latent policy trained in (a).}
\label{fig:overview}
\end{figure*}

We focus on learning behavior correspondence across robots of different embodiments, including different numbers of joints, kinematics, and dynamics. For example, we consider training a pick-and-place policy for a 7 degrees of freedom (DoF) Panda arm and transferring it to a 6 DoF xArm6 robot. 
Our approach is divided into three stages as shown in Fig~\ref{fig:overview}: (i) source domain policy learning, (ii) target domain latent alignment, and (iii) target domain deployment. In the first stage, we project the source states and actions to an embodiment-independent latent space that is used later to relate to the target states and actions. We train encoders and decoders for the latent space projection simultaneously with a latent-space policy using loss functions measuring \emph{task performance}, \emph{latent dynamics consistency}, and encoder-decoder ability for \emph{reconstruction} of the original source states and actions. In the second stage, we train target encoders and decoders that align a new target robot to the same latent space obtained in the first stage. We optimize an \emph{adversarial training objective} such that the latent distribution of target samples matches that of the source samples. Additionally, we introduce a \emph{cycle consistency loss} such that a state-action sample from one domain, projected to the other domain through the latent space, is close to itself when projected back to its original domain. The cycle consistency constraint can leverage unpaired, unaligned, randomly-collected data from the two domains allowing us to relax the strong assumption that paired correspondences between the two domains are available. In the third stage, we compose the target encoder and decoder functions from (ii) with the latent policy (i) to obtain a target-domain policy. This zero-shot target policy does not require access to a target-domain reward function or expert demonstrations, nor does it optimize the RL objective in the target domain. In summary, this paper makes the following contributions.

\begin{itemize}
    \item We develop a novel approach for cross-embodiment robot manipulation skill transfer. Our key idea is to train a policy in a latent space that facilitates alignment with robots with different state and action spaces.

    \item We enable transfer without policy retraining or fine-tuning by aligning the target domain to the latent space via adversarial training and cycle consistency with unpaired, unaligned, randomly-collected target data.
    
    \item We demonstrate that policies for different manipulation tasks trained on a simulated Panda robot can be transferred to simulated Sawyer and real xArm6 robots. 
\end{itemize}

    
    



\section{Related Work}
\label{sec:related_work}

Learning invariant features has shown promise for solving RL and domain transfer tasks. Guo et al. \cite{Guo2020Bootstrap} and Hafner et al. \cite{Hafner2019PlaNet} independently propose to learn a latent recurrent state model from pixel observations and bootstrap latent dynamics to learn latent predictions of future observations. Pari et al. \cite{Pari2021Surprising} decouple latent representation learning from robot behavior learning. First, they use offline visual data to train a bootstrap-style self-supervised model \cite{Grill2020BYOL} where the encoder learns to project augmented versions of the same image to similar latent representations. During inference, the latent input is compared against the nearest neighbor latents from the demonstrations to query the action.

Learning visual and state representations is useful for training better robot manipulation policies or aligning with human demonstrations. Zhang et al. \cite{Zhang2018Deep} use teleoperation to align human demonstrations with robot arms for imitation learning. Wang et al. \cite{Wang2019VPA} consider a visual manipulation planning problem where a causal InfoGAN model \cite{Kurutach2018CIGAN} generates future visual observations with latent planning and a learned inverse dynamics model tracks the predicted observation sequence. Nair at el. \cite{Nair2022R3M} pre-train latent visual representations using time-contrastive learning \cite{Sermanet2018Time} and video-language alignment \cite{Nair2022Learning} to be used in robot manipulation tasks. Das et al. \cite{Das2021MBIRL} use visual keypoints as latent features to imitate human demonstrations for robot manipulation tasks.

Transfer learning leverages the knowledge learned on one task to finetune on a second related task. Zhang et al. \cite{Zhang2020Bisim} learn invariant representations through bisimulation metrics \cite{Ferns2014Bisim} where states are considered similar if their immediate rewards and state distributions are similar under the same action sequence. Wulfmeier et al. \cite{Wulfmeier2017Mutual} finetune a source policy on a target robot of the same type but with different dynamics by encouraging a similar state distributions. Zakka et al. \cite{Zakka2022XIRL} use a temporal consistency constraint with paired source and target samples for cross-embodiment imitation learning. Hejna et al. \cite{Hejna2020Morph} achieve cross-morphology transfer by finetuning hierarchical policies with a Kullback–Leibler divergence constraint. Zhang et al. \cite{Zhang2021CrossDomain} learn a direct state and action correspondence between source and target domains. However, they do not construct a latent space so the correspondence has to be re-trained for each new domain. Stadie et al. \cite{Stadie2017Third} consider imitation learning under viewpoint mismatch by training viewpoint-invariant features.
Kim et al. \cite{Kim2020DAIL} consider matching state distributions via generative adversarial networks across domains in an imitation learning setting. In contrast, our work does not require expert demonstrations in target domains. Yin et al. \cite{Yin2022CrossDomain} learn a latent invariant representation for a robot with different physical parameters (e.g. link length). However, the method cannot be applied if the source and target robots have different morphology (e.g., different number of links). Yoneda et al.~\cite{Yoneda22ILA} align latent features from source and target samples with adversarial training and dynamics consistency constraint. While their approach only considers visual adaptation for the same robot, we consider a more general setting of aligning robots of different embodiments. 

Our work is inspired by cycle consistency techniques. CycleGAN \cite{Zhu2017CycleGAN} uses a cycle consistency loss for unpaired image-to-image translation. Rao et al. \cite{Rao2020RLCycleGAN} extend CycleGAN to reinforcement learning with an additional Q-function supervision loss and show sim2real transfer for vision-based grasping. Recently, \cite{Bousmalis2023RoboCat} proposed a transformer model with tokenized embeddings to encode visual observations and generalize across different robot embodiments. 


\section{Problem Formulation}
\label{sec:problem_formulation}

A Markov decision process (MDP) $\calM =\crl{\calS, \calA, r, T, \gamma}$ consists of a continuous state space $\calS$, a continuous action space $\calA$, a reward function $r: \calS \times \calA \rightarrow \bbR$, a probabilistic transition function $T: \calS \times \calA \times \calS \rightarrow \brl{0, 1}$, and a discount factor $\gamma \in \brl{0,1 }$. We consider a source MDP $\calM^s = \crl{\calS^s, \calA^s, r^s, T^s, \gamma}$ and a target MDP $\calM^t = \crl{\calS^t, \calA^t, r^t, T^t, \gamma}$ with different state and action spaces. We aim to align the source and the target domains by defining a latent-space MDP $\calM^z = \crl{\calS^z, \calA^z, r^z, T^z, \gamma}$ and encoder and decoder functions that relate the source and target domains to the latent-space MDP. We introduce a state encoder $F^s: \calS^s \rightarrow \calS^z$ and an action encoder $G^s: \calS^s \times \calA^s \rightarrow \calA^z$ to map source state-action pairs to the latent MDP. We also introduce decoders $\tdF^s: \calS^z \rightarrow \calS^s$ and $\tdG^s: \calS^s \times \calA^z \rightarrow \calA^s$ to map latent state-action pairs back. Similarly, we define state-action encoders $F^t$, $G^t$ and decoders $\tdF^t$, $\tdG^t$ between the target MDP and the latent MDP. 

Assume that random transitions $\calD^s = \crl{(\bfs^s_k, \bfa^s_k, \bfs^s_{k+1})}$ and $\calD^t = \crl{(\bfs^t_k, \bfa^t_k, \bfs^t_{k+1})}$ are available from the source and target domains. Our goal is to learn the state-action encoders and decoders $F^{\crl{s,t}}, \tdF^{\crl{s,t}}, G^{\crl{s,t}}, \tdG^{\crl{s,t}}$ such that a source policy parameterized through the latent space, $\pi^s(\bfs^s) = \tdG^s(\bfs^s, \pi^z(F^s(\bfs^s)))$, can be transferred to the target domain by keeping the latent policy $\pi^z$ fixed and only replacing the embedding functions, i.e.,  $\pi^t(\bfs^t) = \tdG^t(\bfs^t, \pi^z(F^t(\bfs^t))$. We consider a deterministic latent policy $\pi^z: \calS^z \rightarrow \calA^z$ for simplicity of presentation. 

In the context of robotics, the encoders and decoders provide a common latent space in which different robot emobodiments are aligned. A trained latent policy can be reused when a new target robot is introduced without learning a new target policy from scratch or requiring new task-specific data.


\section{Cross Embodiment Representation Alignment}
\label{sec:approach}

\begin{figure*}[t]
\centering
\includegraphics[width=.95\textwidth]{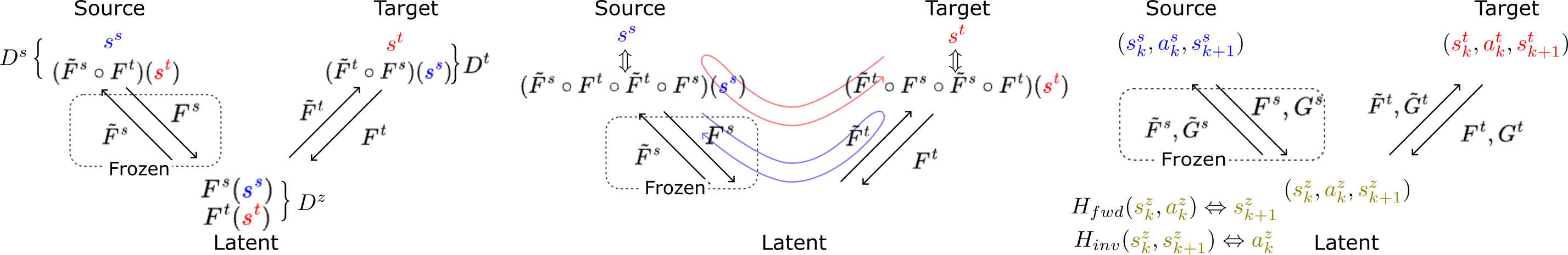}
\caption{Overview of target domain alignment losses: (left) the adversarial loss ensures that the state-action distributions in the source and target domain match, (middle) the cycle consistency loss regularizes state-action samples to be close to themselves when translated to the other domain and back, (right) the latent dynamics loss enforces consistent forward and inverse latent transitions.}
\label{fig:alignment_losses}
\end{figure*}

In this section, we present an approach to train the source-domain encoders and decoders as well as a latent space policy to achieve a desired task. Next, we show how to train target-domain encoders and decoders by aligning target-domain samples to the latent space constructed in the first stage. This allows transferring the latent space policy learned in the source domain without requiring additional task-specific training data in the target domain.


\subsection{Latent Policy Training with Source Domain Alignment}
\label{sec:source_policy}

Training a policy on a given robot system directly does not help build a representation that allows transferring the learned skill to a new system. To enable skill transfer, we parameterize a source-domain policy in $\calM^s$ via a latent policy in $\calM^z$ using a state encoder $F^s$ and an action decoder $\tdG^s$: $\pi^s(\bfs^s) = \tdG^s(\bfs^s, \pi^z(F^s(\bfs^s)))$. Instead of directly predicting a source action from a source state, we project the source state $\bfs^s$ to a latent state $\bfs^z = F^s(\bfs^s)$, use the latent policy to predict a latent action $\bfa^z = \pi^z(\bfs^z)$, and project the latent action back to a source action $\bfa^s = \tdG^s(\bfs^s, \bfa^z)$. 

We train the encoders, decoders, and latent policy in the source domain simultaneously using three loss functions: \emph{task}, \emph{latent dynamics}, and \emph{reconstruction}. The task loss captures the objective that the policy needs to optimize. Since the latent-space MDP $\calM^z$ does not correspond to a real system, the latent dynamics loss ensures that the latent transitions are consistent forward and backward in time. The reconstruction loss ensures that the state-action encoders and decoders are inverses of each other. These source-domain training loss functions are described in detail below.


\textbf{Task Loss.}
The first objective is to optimize the source-domain encoders, decoders, and latent policy to perform a desired task. We optimize the state encoder $F^s$ and the action decoder $\tdG^s$ jointly with the latent policy $\pi^z$ using a deterministic policy gradient algorithm, e.g., TD3 \cite{policy_gradient_theorem, Lillicrap2015DDPG, Fujimoto2018TD3}. Recall that deterministic policy gradient algorithms minimize expected cumulative cost (or equivalently maximize expected cumulative reward) of a parameterized policy $\pi_\theta^s$:
\begin{equation}
    \calL_{task}(\theta) = -\bbE_{\pi_\theta^s} \biggl[\sum_{k=0}^\infty \gamma^k r(\bfs_k, \bfa_k)\biggr].
\end{equation}
The gradient of the task loss is:
\begin{equation}
\label{eq:policy_gradient}
\nabla_\theta \calL_{task}(\theta) = -\bbE \brl{\nabla_\theta \pi_{\theta}^s (\bfs) \nabla_{\bfa} Q^{\pi_\theta^s}(\bfs, \bfa)\vert_{\bfa=\pi_\theta^s(\bfs)}},
\end{equation}
where the action-value (Q) function of $\pi_\theta^s$ is 
\begin{equation*}
Q^{\pi_\theta^s}\!(\bfs, \bfa) = \bbE_{\bfs\sim T_{\pi_\theta}, \bfa \sim \pi_\theta^s} \biggl[\sum_{k=0}^\infty \gamma^k r(\bfs_k, \bfa_k) \vert \bfs_0 = \bfs, \bfa_0 = \bfa \biggr]\!,
\end{equation*}
i.e., the expected sum of rewards when choosing action $\bfa$ in state $\bfs$ and following $\pi_\theta^s$ afterwards. In our approach, since the policy is a composition of the encoder $F^s$, decoder $\tdG^s$ and latent policy $\pi^z$, the policy gradient in \eqref{eq:policy_gradient} is backpropagated through each component to update its parameters.


\textbf{Latent Dynamics Loss.}
The task loss above optimizes the latent state and action representations to achieve good task performance but does not ensure that the learned representations correspond to consistent latent-space transitions. We introduce a self-supervision signal based on latent dynamics prediction \cite{Hansen2020PAD} to learn consistent forward and inverse dynamics models $T^z: \calS^z \times \calA^z \rightarrow \calS^z$ and $\tdT^z: \calS^z \times \calS^z \rightarrow \calA^z$ in the latent space:
\begin{align}
\label{eq:latent_dynamics_loss}
&\calL_{dyn}(T^z, \tdT^z, F^s, G^s) = \bbE_{(\bfs_k^s, \bfa_k^s, \bfs^s_{k+1})\sim D^s} \nonumber\\
&\brl{\norm{T^z(\bfs_k^z, \bfa_k^z) - \bfs^z_{k+1}}_2^2 + \norm{\tdT^z(\bfs^z_k, \bfs^z_{k+1}) - \bfa^z_k}_2^2} 
\end{align}
where $\bfs_k^z = F^s(\bfs_k^s)$ and $\bfa_k^z = G^s(\bfs_k^s, \bfa_k^s)$. We use deterministic latent dynamics for ease of implementation but a stochastic dynamics model can also be considered \cite{lee2020slac}. The dynamics loss is optimized simultaneously with the task loss.

\textbf{Reconstruction Loss.} 
Finally, we introduce a reconstruction loss that ensures that the state and action encoders and decoders are consistent in the sense of being inverses of each other. Reconstruction losses are not always used with visual observations to avoid reconstructing distractions and noise but, if the source domain captures robot joint configuration, this information should be retained accurately in the latent space. Therefore, we require that $F^s$ and $\tdF^s$, as well as $G^s$ and $\tdG^s$, are inverse mappings of each other:
\begin{align}
\label{eq:recon_loss}
&\calL_{rec}(F^s, \tdF^s, G^s, \tdG^s) = \bbE_{(\bfs^s_k, \bfa^s_k)\sim D^s} \\
&\brl{\norm{\tdF^s(F^s(\bfs_k^s)) - \bfs^s_k}_2^2 + \norm{\tdG^s(\bfs^s_k, G(\bfs^s_k, \bfa^s_k)) - \bfa^s_k}_2^2}.\nonumber
\end{align}

\textbf{Training Algorithm.} 
The pseudo code for learning a source-domain policy with latent-space parametrization is shown in Alg.~\ref{alg:source_policy}. We first initialize a replay buffer $\calD$ containing the random source transitions $\calD^s$. We update the $F^s, \pi^z, \tdG^s$ with the task policy gradient in \eqref{eq:policy_gradient} and use the same minibatch samples to optimize the latent dynamics $T^z, \tdT^z$ with the dynamics loss in \eqref{eq:latent_dynamics_loss} and to enforce the encoder-decoder reconstruction loss in \eqref{eq:recon_loss}.


\begin{algorithm}[t]
\caption{Source domain policy training}
\label{alg:source_policy}
\small
\begin{algorithmic}[1]
\State Initialize replay buffer $\calD$ with source samples $\calD^s$.
\Loop
    \State Select action $\bfa = \tdG^s(\bfs^s, \pi^z(F^s(\bfs^s)) + \epsilon$ with exploration noise $\epsilon \sim \calN(0, \sigma^2)$
    \State Observe reward $r$, next state $\bfs'$ and store $(\bfs, \bfa, r, \bfs')$ in $\calD$
    \State Sample a minibatch $\crl{(\bfs^s_k, \bfa^s_k, r^s_k, \bfs^s_{k+1})}$ from $\calD$
    \State Update $\tdG^s, \pi^z, F^s$ with deterministic policy gradient in \eqref{eq:policy_gradient}
    \State Update latent dynamics $T^z, \tdT^z$ and encoders $F^s, G^s$ with latent dynamics loss in \eqref{eq:latent_dynamics_loss}
    \State Update encoders $F^s, G^s$ and decoders $\tdF^s, \tdG^s$ with reconstruction loss in \eqref{eq:recon_loss}
\EndLoop
\State {\bfseries Output:} Source encoders and decoders $F^s, \tdF^s, G^s, \tdG^s$, latent dynamics $T^z, \tdT^z$, latent policy $\pi^z$.
\end{algorithmic}
\end{algorithm}

\subsection{Target Domain Alignment}
\label{sec:alignment}

Next, we consider aligning a target domain to the latent domain constructed during policy training in the source domain. This allows us to construct a target policy $\pi^t(\bfs^t) = \tdG^t(\bfs^t, \pi^z(F^t(\bfs^t))$, in which the task-dependent latent policy $\pi^z$ is fixed (after the source-domain training in Sec.~\ref{sec:source_policy}) and only the task-independent target-domain encoders and decoders need to be trained. We use three loss functions to train the target-domain encoders and decoders, as shown in Fig.~\ref{fig:alignment_losses}: \emph{adversarial}, \emph{cycle consistency}, and \emph{latent dynamics}. The adversarial loss ensures that state-action distributions obtained in the latent domain from the source encoders and from the target encoders are similar. The cycle consistency loss ensures that translating any state-action pair from the source to the target domain and back recovers the same state-action pair. This loss is critical for training the encoders and decoders with unpaired samples in the two domains. The latent dynamics consistency loss is the same as in the source training stage and ensures that the encoded target samples follow the same dynamics in the latent space.




\textbf{Adversarial Loss.}
The encoders and decoders learned in the source domain fix the latent distribution of the source-domain transitions. We aim to train the target-domain encoders and decoders $F^t, \tdF^t, G^t, \tdG^t$ such that the latent distribution of the target-domain transitions $\calD^t$ matches the latent distribution of the source-domain transitions $\calD^s$. We consider an adversarial learning approach, where a discriminator $D^z: \calS^z \times \calA^z \rightarrow [0, 1]$ tries to distinguish between $\prl{F^s(\bfs^s), G^s(\bfs^s, \bfa^s)}$ from the source domain and $\prl{F^t(\bfs^t), G^t(\bfs^t, \bfa^t)}$ from the target domain. The distribution of $\prl{F^s(\bfs^s), G^s(\bfs^s, \bfa^s)}$ is fixed because $F^s, G^s$ are trained in Sec.\ref{sec:source_policy} and are frozen during target domain alignment. The target domain encoders $F^t, G^t$ act as a generator aiming to synthesize latent state-action pairs $F^t(\bfs^t), G^t(\bfs^t, \bfa^t)$ that are indistinguishable from the source latents. The adversarial loss in the latent space is defined as: 
\begin{align}
&\max_{D^z}\min_{F^t, G^t}\calL_{gan}^z(F^t, G^t, D^z) = \nonumber\\
&\quad\bbE_{\prl{\bfs^s, \bfa^s}\sim \calD^s} \brl{\log D^z(F^s(\bfs^s), G^s(\bfs^s, \bfa^s))} + \nonumber\\
&\quad\bbE_{\prl{\bfs^t, \bfa^t}\sim \calD^t} \brl{\log (1 - D^z(F^t(\bfs^t), G^t(\bfs^t, \bfa^t)))}.
\end{align}
We may also match the translated state-action distributions in the source and in the target domains. For example, in the target domain, a state-action pair translated from the source domain is $\brbfs^t = \tdF^t(F^s(\bfs^s))$ and $\brbfa^t = \tdG^t(\brbfs^t, G^s(\bfs^s, \bfa^s))$. With another discriminator $D^t: \calS^t \times \calA^t \rightarrow [0, 1]$ which distinguishes real target pairs $\prl{\bfs^t, \bfa^t}$ from generated ones $\prl{\brbfs^t, \brbfa^t}$, the adversarial loss in the target domain is:
\begin{align}
&\max_{D^t}\min_{\tdF^t, \tdG^t}\calL_{gan}^t(\tdF^t, \tdG^t, D^t) = \nonumber \\ 
&\resizebox{.9\hsize}{!}{$\bbE_{(\bfs^t, \bfa^t)\sim \calD^t} \brl{\log D^t(\bfs^t, \bfa^t)} + \bbE_{(\bfs^s, \bfa^s)\sim \calD^s} \brl{\log (1-D^t(\brbfs^t, \brbfa^t)}.$}
\end{align}
Similarly, we can construct a source-domain discriminator $D^s$ which distinguishes the translated target distribution in the source domain:
\begin{align}
&\max_{D^s}\min_{F^t, G^t}\calL_{gan}^s(F^t, G^t, D^s) = \nonumber \\ 
&\resizebox{.9\hsize}{!}{$\bbE_{(\bfs^s, \bfa^s)\sim \calD^s} \brl{\log D^s(\bfs^s, \bfa^s)} + \bbE_{(\bfs^t, \bfa^t)\sim \calD^t} \brl{\log (1-D^s(\brbfs^s, \brbfa^s)}$},
\end{align}
where $\brbfs^s = \tdF^s(F^t(\bfs^t))$ and $\brbfa^s = \tdG^s(\brbfs^s, G^t(\bfs^t, \bfa^t))$. Combining the adversarial objectives in the latent, source, and target domains leads to the complete adversarial loss: 
\begin{align}
\label{eq:adv_loss}
\max_{D^z,D^s,D^t}\min_{F^t, \tdF^t, G^t, \tdG^t} \calL_{gan}^z + \calL_{gan}^s + \calL_{gan}^t.
\end{align}

\textbf{Cycle Consistency Loss.}
Inspired by CycleGAN \cite{Zhu2017CycleGAN}, we construct a cycle consistency loss such that translating a state-action pair from one domain to the other and back recovers the same state-action pair. This loss leverages unpaired samples since it only requires samples from one domain, obviating the need for paired samples from both domains. Specifically, if we have a translated target state $\brbfs^t = \tdF^t(F^s(\bfs^s))$ from a source state, the reconstructed source state from it should be close to itself, i.e., $\bar{\brbfs}^s = \tdF^s(F^t(\brbfs^t)) \approx \bfs^s$. The cycle consistency objective also applies to translated actions, i.e., $\bar{\brbfa}^s = \tdG^s(\tdF^s(F^t(\brbfs^t)), G^t(\brbfs^t, \brbfa^t)) \approx \bfa^s$. The full cycle consistency loss for both domains is:
\begin{align}
\label{eq:cycle_loss}
&\resizebox{.95\hsize}{!}{$\calL_{cyc}(F^t, \tdF^t, G^t, \tdG^t) = \bbE_{\prl{\bfs^s, \bfa^s}\sim \calD^s} \brl{\norm{\bar{\brbfs}^s - \bfs^s}_1 + \norm{\bar{\brbfa}^s - \bfa^s}_1}$} \nonumber \\
&\quad + \bbE_{\prl{\bfs^t, \bfa^t}\sim \calD^t} \brl{\norm{\bar{\brbfs}^t - \bfs^t}_1 + \norm{\bar{\brbfa}^t - \bfa^t}_1}.
\end{align}

\textbf{Latent Dynamics Loss.}
In the previous section, we trained latent dynamics models $T^z$ and $\tdT^z$ using source samples. During alignment, we train target encoders such that the latent dynamics consistency still holds for target samples: 
\begin{align}
\label{eq:dyn_loss}
&\calL_{dyn,t}(F^t, G^t) = \bbE_{(\bfs_k^t, \bfa_k^t, \bfs^t_{k+1})\sim D^t} \nonumber\\
&\quad\brl{\norm{T^z(\bfs_k^z, \bfa_k^z) - \bfs^z_{k+1}}_2^2 + \norm{\tdT^z(\bfs^z_k, \bfs^z_{k+1}) - \bfa^z_k}_2^2}, 
\end{align}
where $\bfs^z_k = F^t(\bfs^t_k)$, $\bfa^z_k = G^t(\bfs^t_k, \bfa^t_k)$. Here $T^z$, $\tdT^z$ are fixed during the target latent alignment stage since we want the target samples to follow the same latent transitions induced by the source domain training in Sec.~\ref{sec:source_policy}.


\textbf{Transfer Algorithm.}
After training the target-domain encoders and decoders $F^t, \tdF^t, G^t, \tdG^t$ with the above losses, the source and target samples $\calD^s, \calD^t$ are aligned to the common latent space $\calM^z$. Alg.~\ref{alg:alignment} summarizes the target-domain alignment procedure. During target deployment, the latent policy $\pi^z$ from Sec.~\ref{sec:source_policy} is reused while the state encoder and action decoder are replaced with $F^t$ and $\tdG^t$ to obtain the target-domain policy $\pi^t(\bfs^t) = \tdG^t(\bfs^t, \pi^z(F^t(\bfs^t)))$. Our approach does not require paired source and target data or reward supervision for target-domain transfer.



\begin{algorithm}[t]
\caption{Target domain policy transfer}
\label{alg:alignment}
\small
\begin{algorithmic}[1]
\State Freeze learned models $F^s, \tdF^s, G^s, \tdG^s, T^z, \tdT^z, \pi^z$ from Alg.~\ref{alg:source_policy}
\Loop
    \State Sample $\crl{(\bfs^s_k, \bfa^s_k, \bfs^s_{k+1})} \!\sim\! D^s$ and $\crl{(\bfs^t_k, \bfa^t_k, \bfs^t_{k+1})} \!\sim\! D^t$
    \State Update discriminators $D^z, D^s, D^t$ by maximizing \eqref{eq:adv_loss}
    \State Update target encoders and decoders $F^t, \tdF^t, G^t, \tdG^t$
 by minimizing \eqref{eq:adv_loss}, \eqref{eq:cycle_loss}, \eqref{eq:dyn_loss}
\EndLoop
\State {\bfseries Output:} Target policy:  $\pi^t(\bfs^t) = \tdG^t(\bfs^t, \pi^z(F^t(\bfs^t)))$
\end{algorithmic}
\end{algorithm}

\begin{figure}[t]
\centering
\includegraphics[width=.98\linewidth]{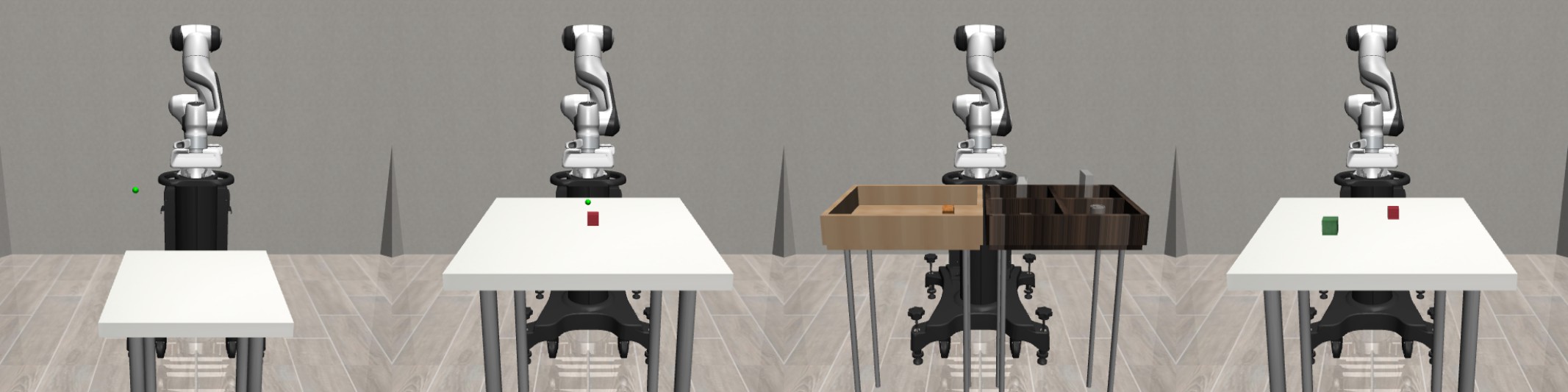}
\caption{Robosuite simulation tasks (from left to right): Reach, Lift, PickPlace and Stack.}
\label{fig:envs}
\end{figure}

\section{Experiments}
\label{sec:experiments}

In this section, we evaluate the ability of our method for policy transfer to simulated and real robot manipulators.

\subsection{Simulation Results}
\label{sec:simulation_results}

We evaluated our policy-transfer approach and conducted ablation studies in four robot manipulation tasks, Reach, Lift, PickPlace, and Stack, in the Robosuite simulation environment \cite{robosuite2020}, illustrated in Fig.~\ref{fig:envs}. Details about the tasks and the simulation setup are presented in Appendix \ref{sec:appendix}.


\textbf{Reach Task.}
In the first experiment, we considered the same Panda robot of different state spaces on the Reach task. Specifically, the source domain was a Panda with a 3-D end-effector position state space and 3-D end-effector position displacement action space. The target domain was the same Panda robot with the same action space but using 7-D joint angles as states. We set the latent-space MDP $\calM^z$ equal to the source-domain MDP $\calM^s$, i.e., $F^s, \tdF^s, G^{s,t}, \tdG^{s,t}$ were identity functions, and learned the target state encoder decoder $F^t, \tdF^t$ only. In this setting, $F^t, \tdF^t$ should theoretically approximate the forward and inverse kinematics between the joint angles and the end-effector position. 


As an evaluation metric, we computed the $\ell_1$-distance between the predicted and ground-truth end-effector positions to assess the state alignment. We also evaluated the RL performance by training a source policy $\pi^s$ using TD3 \cite{Fujimoto2018TD3} ($\pi^s = \pi^z$ since $F^s = \tdG^s = I$) and transferring it to a target policy $\pi^t(\bfs^t) = \pi^z(F^t(\bfs^t))$ (since $\tdG^t = I$).


We compared with the following baselines: (i) invariance through latent alignment (ILA) \cite{Yoneda22ILA} which performs state alignment in the source domain and does not require cycle consistency; (ii) our model trained without a latent dynamics loss; (iii) a strongly supervised model that is trained on paired end-effector and joint-angle data; (iv) an oracle model obtained by training an RL policy directly in the target domain; (v) a random policy in the target domain. Table \ref{tb:reach_results} shows that the cycle consistency and dynamics consistency losses are important for improving the performance of our model over ILA \cite{Yoneda22ILA}. The strongly supervised model is marginally better than ours even though it uses paired source and target data while our model uses unpaired data.

\begin{table}[t]
\centering
\caption{Transfer learning on Reach task with Panda end-effector space source and Panda joint space target. The results are averaged over 5 runs. Lower is better for the $\ell_1$ error (cm). Higher is better for the RL score. Our model performs better using a latent dynamics loss, and outperforms ILA \cite{Yoneda22ILA} which does not use cycle consistency. The strong supervision model requires paired training data. The oracle is an RL policy trained in the target domain.}
\resizebox{\linewidth}{!}{\begin{tabular}{*7c}
\Xhline{2\arrayrulewidth}
& Ours & \makecell{Ours\\w/o dyn.} & ILA \cite{Yoneda22ILA} & \makecell{Strong\\supervision} & Oracle & Random \\
\hline
$\ell_1$ error & $0.7 \pm 0.1$ & $4.7 \pm 0.2$ & $3.9 \pm 0.3$ & $0.3 \pm 0.1$ & - & - \\
RL score & $163 \pm 10$ & $41 \pm 27$ & $53 \pm 27$ & $166 \pm 7$ & $172 \pm 11$ & $13 \pm 8$ \\
\Xhline{2\arrayrulewidth}
\end{tabular}}
\label{tb:reach_results}
\end{table}



\begin{figure}[t]
\centering
\includegraphics[width=\linewidth]{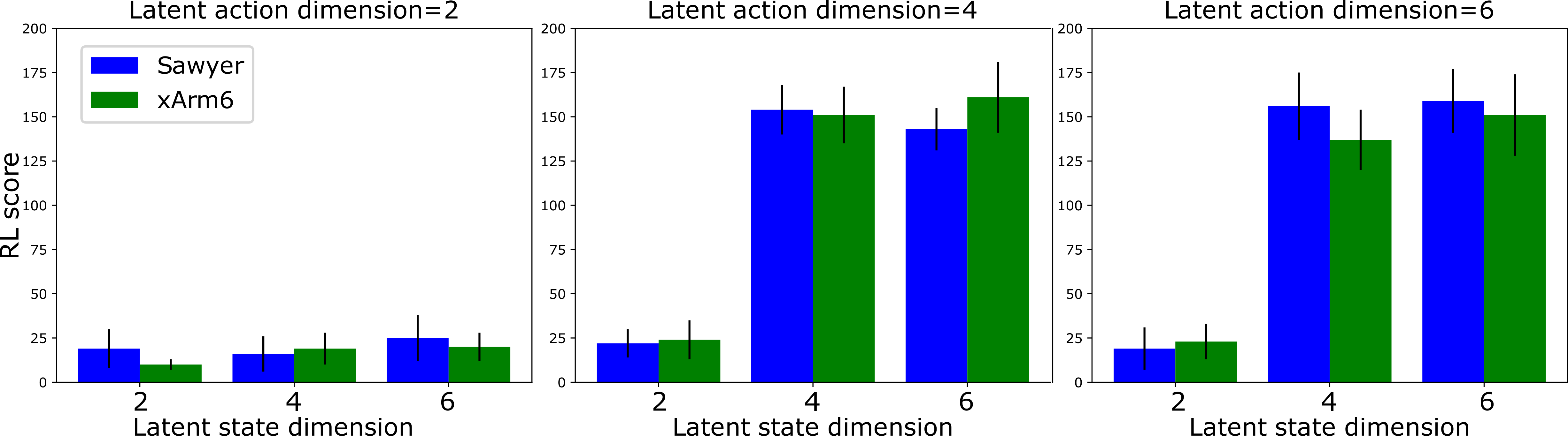}
\caption{Ablation on latent state and action dimensions for policy transfer from Panda to Sawyer and xArm6 robots on the Reach task. The lowest state and action dimensions with reasonable performance are 4.}
\label{fig:dim_ablation}
\end{figure}

\textbf{Reach Task Ablation.}
Next, we performed an ablation study on the dimensions of the latent state and action spaces for the Reach task. We used a 7 DoF Panda arm as source and a 7 DoF Sawyer arm and a 6 DoF xArm6 arm as target domains. The source and target state dimensions, represented in sine and cosine functions of the joint angles, were 14 for the Panda and the Sawyer and 12 for the xArm6. Joint velocities were used as the action space for each robot with dimension equal to the DoF. Fig. \ref{fig:dim_ablation} shows the performance of a control policy transferred from the Panda source domain to the Sawyer and xArm6 target domains with different latent state $\bfs^z$ and action $\bfa^z$ dimensions. We found that the transfer performance drops significantly when the state or action dimensions are too small. This makes sense as end-effector position control takes place in 3D space and latent trajectories in a lower dimensional space cannot recover the 3D motions. In the subsequent experiments, we choose the latent state and action dimensions to both be 4, corresponding to the smallest dimensions that achieve good performance.


\textbf{Lift, PickPlace, Stack Tasks.}
Following a similar setup as in the Reach ablation study, we used joint velocity control to transfer control policies for Lift, PickPlace, and Stack tasks from a Panda arm source domain to Sawyer and xArm6 target domains in the Robosuite simulator \cite{robosuite2020}. The state variables included robot joint angles, gripper width, gripper touch signal, and object and object goal positions. The experiment settings are described in detail in Appendix \ref{sec:experiment_settings}. Fig. \ref{fig:align_experiment} shows a Lift task example of the transferred policy from Panda to Sawyer and xArm6 robots. Table \ref{tb:transfer_results} shows quantitatively that our model learns a meaningful mapping between robots of different embodiments. However, manipulation tasks require precise alignment to correctly grasp objects and, sometimes, the transferred policy cannot complete the tasks successfully.


\begin{table}[t]
\centering
\caption{RL reward for a source policy trained on Panda and transferred to Sawyer and xArm6 robots. The reward of an oracle policy trained directly on the target robot is shown in parenthesis.}
\label{tb:transfer_results}
\resizebox{\linewidth}{!}{
\begin{tabular}{*4c}
\Xhline{2\arrayrulewidth}
Task & Lift & PickPlace & Stack \\
\hline
Sawyer & $122 \pm 41$ ($181 \pm 4$) & $70\pm 35$ ($86 \pm 15$) & $85 \pm 40$ ($121 \pm 18$) \\
xArm6 & $132 \pm 23$ ($171 \pm 7$) & $74 \pm 36$ ($84 \pm 9$) & $78 \pm 49$ ($116 \pm 23$) \\
\Xhline{2\arrayrulewidth}
\end{tabular}}
\end{table}

\begin{figure*}[t]
\centering
\includegraphics[width=0.95\linewidth]{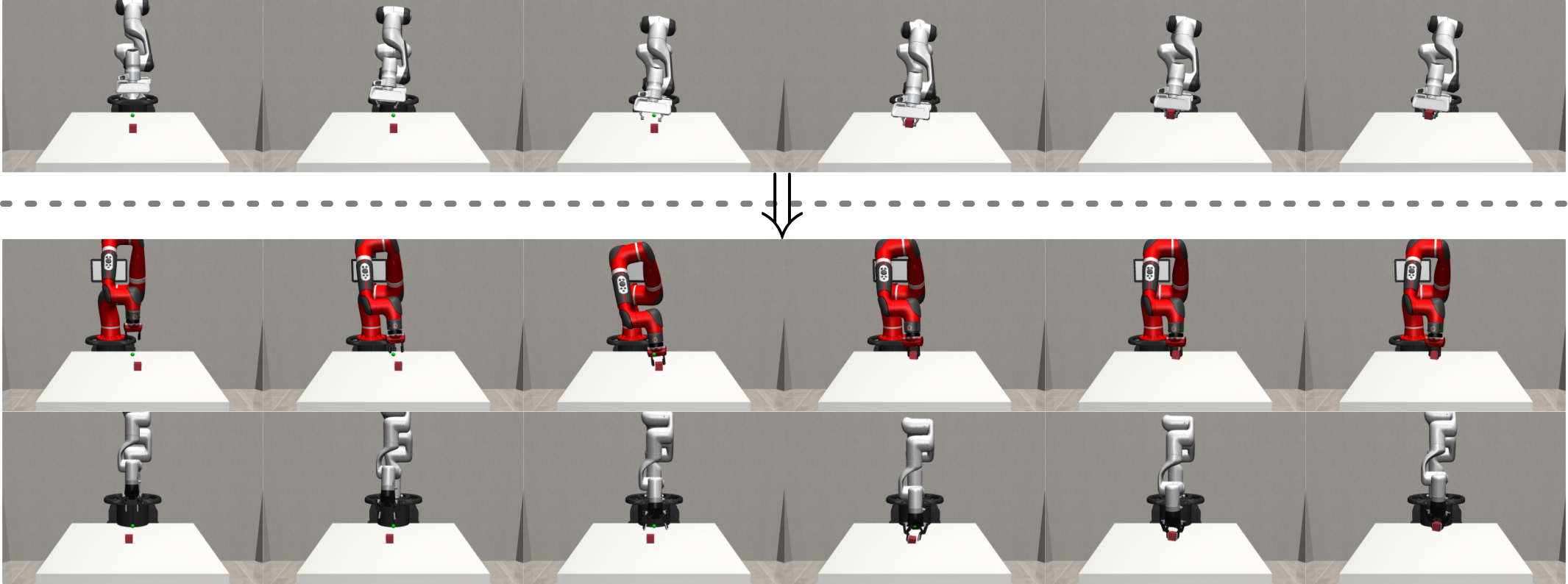}
\caption{Examples of transferring Panda robot policy (top row) to Sawyer robot (middle row) and xArm6 robot (bottom row) for the Lift task in the Robosuite simulator. We learned encoders to map the source robot states and actions to a latent space and simultaneously a latent policy to perform the Lift task. The latent space can be used to align different types of target robots (Sawyer or xArm6) and successfully transfer the learned policy without finetuning with task-specific data in the target domain.}
\label{fig:align_experiment}
\end{figure*}

\subsection{Robot Experiments}

\begin{figure*}[t]
\begin{minipage}[c]{0.265\linewidth}
    \centering
    \includegraphics[width=\linewidth]{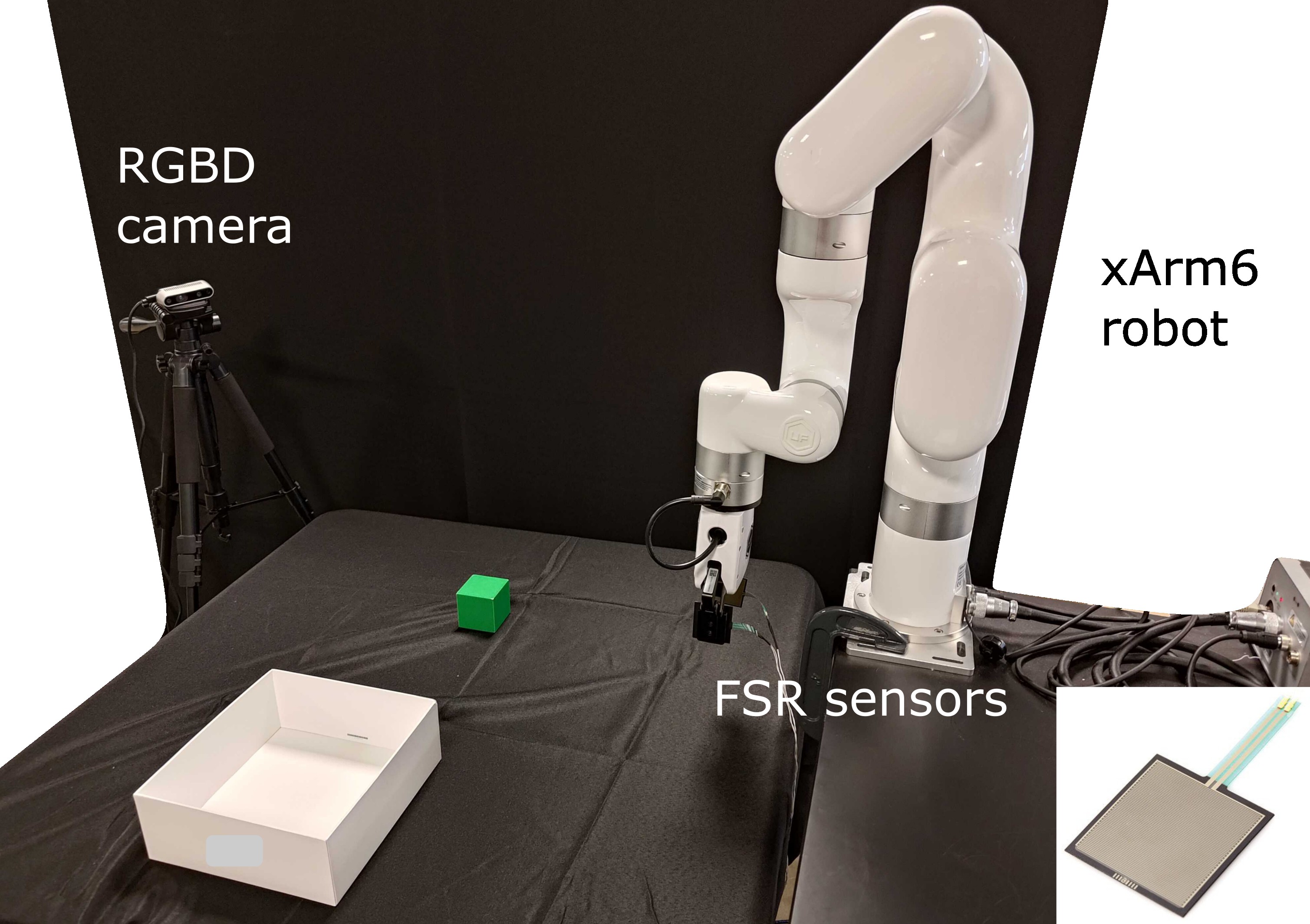}
    \caption{Real-world experiment setup with an xArm6 robot and an RGBD camera for object position estimation. Force sensing resistor (FSR) sensors (bottom right) are attached to grippers to obtain pressure signals.}
    \label{fig:real_setup}
\end{minipage}\hfill
\begin{minipage}[c]{0.72\linewidth}
    \centering
    \includegraphics[width=\linewidth]{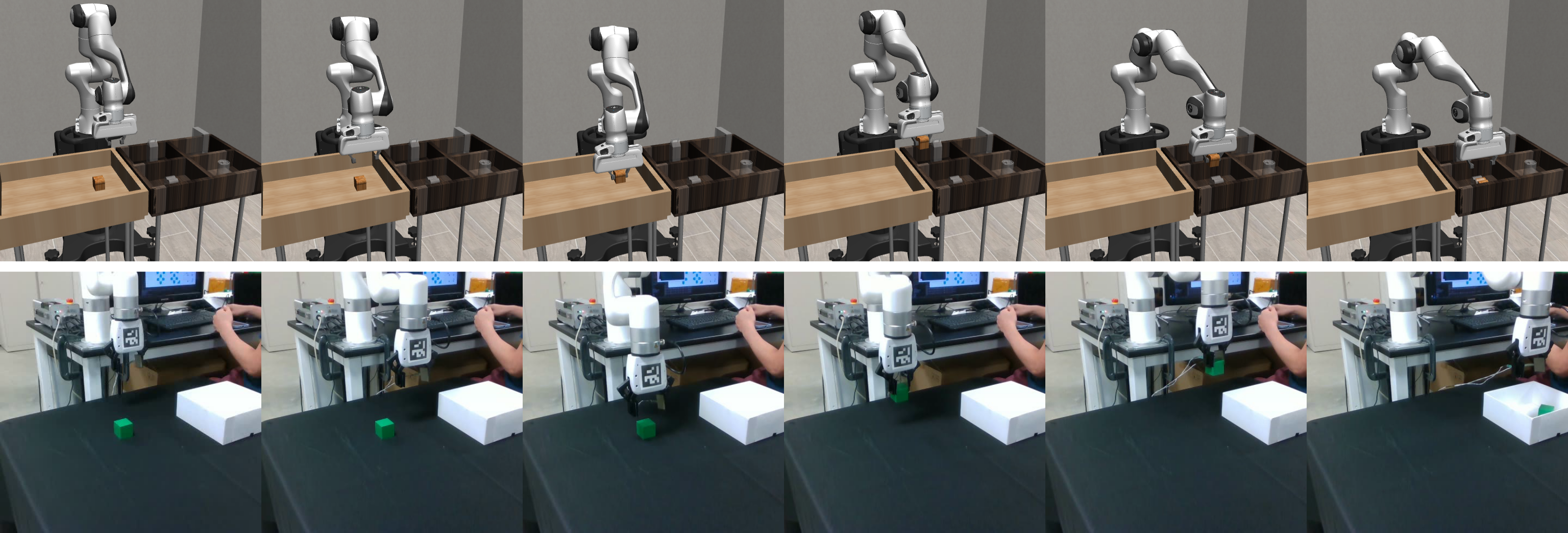}
    \caption{Sim to real transfer for PickPlace task. The source policy is trained with behavior cloning in simulation on a Panda robot (top row) and transferred to a real xArm6 robot (bottom row).}
    \label{fig:sim2real_Panda_xArm6_pickplace}
\end{minipage}
\end{figure*}

We evaluated our model's capability for sim-to-real skill transfer for Lift and PickPlace tasks. We used a simulated Panda arm as the source robot and a real xArm6 arm (setup shown in Fig.~\ref{fig:real_setup}) as the target robot. In either the source or target domain, the state space consists of joint angles, gripper width, gripper touch signal, and object and object goal positions and the action space is the joint velocity controls. The source-domain policy was trained using behavioral cloning \cite{bain1995framework, ross2011reduction} on 100 human demonstrated trajectories in each task, which helps avoid jerky motions when transferring the control policy to the real robot. Since ground-truth object information is not available in the real Lift and PickPlace experiments, we used an RGBD camera to estimate the object position in real time. Details about the object tracking are described in Appendix \ref{sec:object_tracking}. We computed the success rate for each task over 10 test episodes as an evaluation metric. The success conditions are defined in Appendix \ref{sec:experiment_settings} for each task. While simulation environments are quite tolerant to collisions, on the real robot collisions cause a safety stop and are counted as failures.

\begin{table}
\centering
\caption{Transfer results of 10 episodes from simulated Panda to real xArm6 with joint velocity control. Our model can successfully transfer from sim to real. The success rate on the real robot is lower due to dropped objects or collisions, which cause a safety stop.}
\label{tb:xArm6_transfer_results}
\begin{tabular}{*4c}
\Xhline{2\arrayrulewidth}
Task & Success & Collision & Drop \\
\hline
Lift & $70\%$ & $20\%$ & $10\%$ \\
PickPlace & $60\%$ & $20\%$ & $20\%$ \\
\Xhline{2\arrayrulewidth}
\end{tabular}
\end{table}

\begin{figure}[t]
\centering
\includegraphics[width=0.98\linewidth, height=1.8cm]{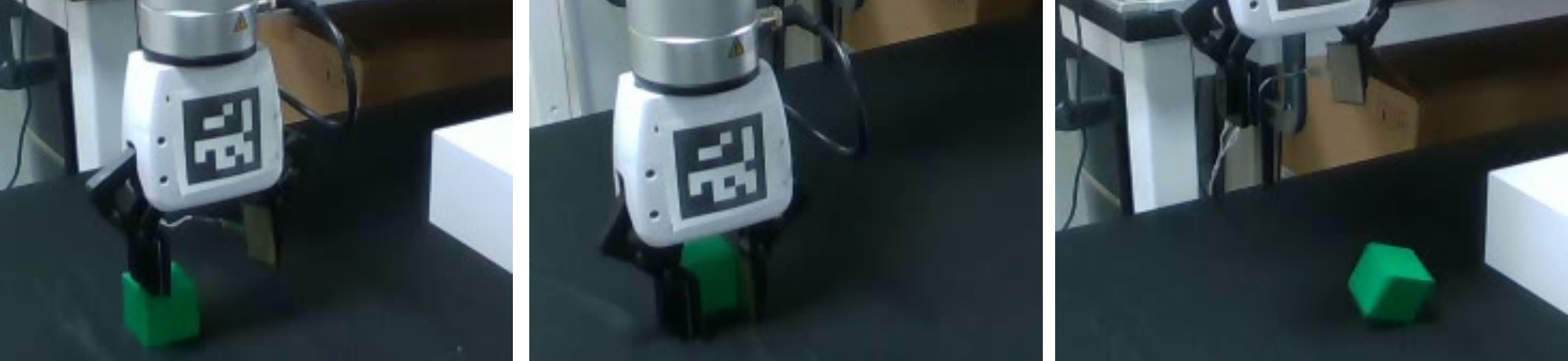}
\caption{Examples of failure modes including cube collisions, table collisions, or permanently dropping the cube.}
\label{fig:xArm6_failures}
\end{figure}

Our method transferred Lift and PickPlace control policies trained on the simulated Panda robot to the real xArm6 robot with reasonable success rate, as shown in Table \ref{tb:xArm6_transfer_results}. Fig. \ref{fig:sim2real_Panda_xArm6_pickplace} shows a successful trajectory of the transferred policy for the PickPlace task, while Fig. \ref{fig:xArm6_failures} shows some failure modes. When the alignment is not perfect, the gripper would collide with the cube or the table. We also observed that the gripper may sometimes drop the cube before it reaches the target. The robot is not able to re-grasp the dropped cube if its landing position is outside that of the training distribution.


\section{Conclusion}
\label{sec:conclusion}

This paper introduced an approach to learn a common latent state and action representation across different manipulators to enable policy transfer. We simultaneously train state and action encoders and decoders to project from source to latent domain, as well as a latent policy using RL. During policy transfer, the target domain encoders and decoders learn to project to the same latent space as the source domain using unpaired data from adversarial training with cycle consistency and latent dynamics constraints. Combining the latent policy with the target-domain encoders and decoders allows the policy to be deployed without accessing the reward function or additional demonstrations in the target domain. Our policy transfer approach performs better than baselines that do not use cycle consistency and latent dynamics constraints in both simulated and real robot experiments.

\section{Appendix}
\label{sec:appendix}

\vspace{-0.15cm}
\subsection{Experiment Settings}
\label{sec:experiment_settings}
We considered Reach, Lift, PickPlace, and Stack tasks from the Robosuite simulator \cite{robosuite2020} as shown in Fig. \ref{fig:envs}. In Reach, the robot end-effector has to reach a target 3D position. In Lift, it lifts a cube to a target position. In PickPlace, it picks up an object and places it in a bin. In Stack, it stacks a red cube on top of a green cube. The state space consists of robot-related states, including sine and cosine functions of joint angles or the gripper end-effector position and gripper opening width, and object-related states, including object and goal positions. The action space consists of joint velocities for joint velocity control or delta positions for end-effector position control, and gripper open or close. For Lift, PickPlace, and Stack tasks, we also used touch sensor signal which was obtained by checking collisions on gripper in simulation and by using force sensing resistors on the real xArm6.  


\vspace{-0.15cm}
\subsection{Model Architecture}
The state and action encoders and decoders $F^{s,t}, \tdF^{s,t},$ $G^{s,t}, \tdG^{s,t}$, latent dynamics functions $T^z, \tdT^z$, and discriminators $D^{s,t,z}$ were implemented as neural networks with hidden layers of size $\brl{256, 256, 256}$. We used ReLU activations for all hidden layers and hyperbolic tangent for the final layer on all models except for discriminators where leaky ReLU activations were used. All models were trained with the Adam optimizer using decay rates $\beta_1=0.9$, $\beta_2=0.999$.

\vspace{-0.15cm}
\subsection{Object Tracking}
\label{sec:object_tracking}
Real-time object tracking was performed with an Intel RealSense D435 stereo camera. We used color-based classification to find the pixel location of object centroids. The color range for the green cube in HSV space was from $\brl{30, 80, 50}$ to $\brl{90, 255, 255}$. The 3D position of the centroid pixel was obtained by querying the corresponding depth value, where post-processing filters on the disparity and time were applied to reduce noise. Finally, we obtained the 3D object position in the robot frame by calibrating camera extrinsic parameters using hand-eye calibration with ArUco markers.


\subsection{Dataset Collection}
We collected random trajectories from Panda, Sawyer, and xArm6 robots in the Robosuite simulator for state-action alignment. Each robot was placed such that the gripper initial position is at $\brl{-0.2, 0, 1.05}$. In each episode, the robot gripper moved in straight line to a randomly sampled position in a 3D rectangular region bounded $\brl{-0.2, -0.25, 0.8}$ to $\brl{0.2, 0.25, 1.2}$ without resetting to initial position. We collected $10000$ episodes of length $200$ for each robot. This sampling strategy covers the robot workspace better than randomly sampling actions at each step. 



\balance
{\small
\bibliographystyle{cls/IEEEtran}
\bibliography{bib/ref.bib}
}

\end{document}